\documentclass[conference]{IEEEtran}
\usepackage[utf8]{inputenc}
\usepackage{graphicx}
\usepackage{float}
\usepackage{amsmath, amssymb}
\usepackage{siunitx}
\usepackage{longtable, tabularx}
\usepackage{caption}
\usepackage{subcaption}
\usepackage{hyperref}
\usepackage[version=4]{mhchem}
\usepackage{pdfpages}
\usepackage{cite}
\usepackage{svg}
\usepackage{balance}
\usepackage{url}

\title{Nonlinear Model Predictive Control via Sequential Convex Programming for Drone-to-Drone Docking}

\IEEEoverridecommandlockouts

\author{
Neeraj Balachandar$^{*}$,
Shriram Hari$^{*}$,
Vishnu R. Unni
\thanks{$^{*}$Equal contribution as co-first authors.} 

\thanks{All authors are with the Department of Mechanical and Aerospace Engineering, Indian Institute of Technology Hyderabad, Telangana, India (e-mail: neerajbalachandar@gmail.com; shriramhari2004@gmail.com; vishnu.runni@gmail.com).}
}

\begin{document}
\maketitle

\begin{abstract}

Autonomous mid-air docking of multirotor vehicles under disturbance-driven target motion poses a constrained nonlinear trajectory optimization challenge. This work formulates the docking task as a finite-horizon optimal control problem based on a reduced-order nonlinear model augmented with disturbance states. The resulting problem is solved using sequential convex programming within a receding-horizon framework to generate dynamically feasible docking trajectories. State estimation with noisy measurements is incorporated to enable robust relative motion prediction, while trajectory execution is validated in a high-fidelity rigid-body MuJoCo simulation environment. 

The proposed framework is evaluated for stationary and constant-velocity target motions, demonstrating reliable convergence to the docking interface while satisfying geometric capture constraints. Quantitatively, the method maintains negligible docking-cone violations and terminal state errors within prescribed tolerances, and achieves consistent, safe docking performance for cone half-angles as low as $10^\circ$. Robust operation is observed for wind disturbance levels up to a standard deviation of $0.5$, while preserving bounded approach velocities and stable control effort. These results demonstrate the effectiveness of the SCP-based trajectory optimization framework for disturbance-robust aerial docking under estimation uncertainty.



\end{abstract}

\vspace{0.3em}





\section{Introduction}
Autonomous docking between moving agents is an important capability for cooperative robotic systems, such as in aerial robotics and multi-UAV coordination. Although rendezvous and docking have been extensively studied and demonstrated in spacecraft missions, extending these principles to aerial vehicles, particularly multirotor drones, introduces distinct challenges. Unlike space-based systems, aerial platforms operate in highly disturbed environments, are underactuated, and subject to aerodynamic interactions, sensing uncertainty, and strict kinodynamic constraints. These factors significantly complicate relative navigation, approach, and physical contact between agents, to achieve reliable docking, particularly when the target drone is non-stationary and undergoing disturbed motion. Recent computer vision techniques such as visual servoing and markerless detection using neural networks have enabled improved target tracking and relative localization \cite{onboardvision}. Nevertheless, improved algorithms are required that enable efficient docking or interaction between drones.


\begin{figure}[!t]
    \centering
    \includegraphics[width=1.0\linewidth]{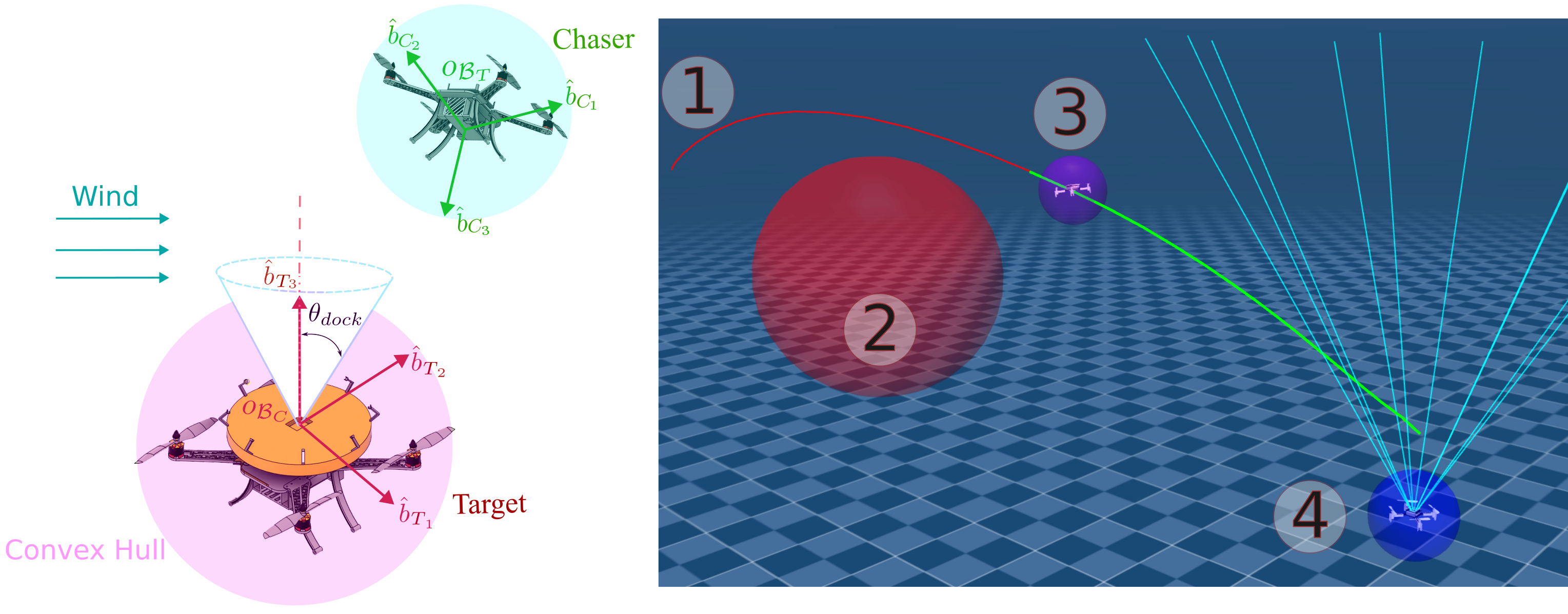}
    \caption{Concept and MuJoCo simulation setup for drone-to-drone docking. The left diagram illustrates the chaser and target reference frames, docking cone geometry, and wind disturbance. The right figure shows the simulation environment with (1) start position, (2) obstacle, (3) chaser drone, and (4) target drone. The green curve denotes the SCP trajectory and the blue rays represent the docking cone. Simulation Video: https://youtu.be/MiMd97Guze4}
    \label{fig:cover}
\end{figure}

Prior works have addressed related problems using linear quadratic regulation for aerial refuelling \cite{lqr} or Model Predictive Control (MPC) for rendezvous and docking of spacecraft \cite{mpc}. For example, \cite{kosari2016optimal} and \cite{xu2020optimal} propose fuzzy-PID-based orientation control and optimal guidance strategies for docking tasks. These approaches typically assume either stationary or well-characterized target dynamics and operate in environments with limited external disturbances and wake interactions. Recent multirotor studies have addressed dynamic path generation for forward-flight docking \cite{docking_multirotor} and airborne docking for multirotor manipulation \cite{multirotor}. However, these works generally assume either stationary or well-characterized targets. They do not explicitly address target motion uncertainty, collision avoidance, and kinodynamic constraints within a unified framework. These limitations motivate optimization-based trajectory planning approaches capable of handling dynamic constraints and safety requirements.





Unlike classical feedback controllers such as PID or LQR, which are effective for stabilization and trajectory tracking under simplified dynamics~\cite{lqr}, the proposed optimization-based approach explicitly handles coupled docking constraints arising from moving-target interception and constrained terminal approach. Compared to standard linear MPC formulations that often rely on simplified geometry or heuristic constraint handling, sequential convex programming (SCP) enables systematic treatment of nonlinear dynamics and nonconvex constraints through iterative convexification. 

Existing SCP-based methods primarily focus on generic trajectory optimization and waypoint navigation~\cite{gusto, waypoint_scp}, while prior aerial docking works address docking system design and trajectory generation under simplified assumptions~\cite{docking_multirotor, multirotor, mpc_docking}.

Here, we build upon the convex soft-capture formulation proposed by Sow et al.~\cite{sow2024convex}, where docking constraints are expressed in a convex form for spacecraft systems. We extend this concept to aerial platforms by incorporating nonlinear multirotor dynamics (as in \cite{lee2010geometric}), actuation limits, and environmental disturbances within a sequential convex programming (SCP) based Nonlinear Model Predictive Control (NMPC) framework.



Accordingly, the drone-to-drone docking task is formulated as a finite-horizon constrained optimal control problem incorporating reduced-order multirotor dynamics, convexified docking constraints, obstacle avoidance, and disturbance-aware target prediction. The resulting SCP-based NMPC framework generates dynamically feasible docking trajectories online under target drift and sensing uncertainty.

Following the structured paradigm commonly adopted in spacecraft docking, the maneuver is divided into two phases: a long-range standoff phase for interception and alignment, followed by a constrained approach phase where geometric docking conditions are enforced. The main contributions of this work are:

\begin{itemize}
    \item \textbf{State estimation and predictive tracking of a distressed target:}  
    Noisy position measurements are used to estimate the states of both the chaser and the target using a Kalman filter. The chaser model includes disturbance terms to account for environmental effects such as wind. The estimated target state is propagated forward and updated at each SCP iteration, allowing the chaser to plan trajectories that remain feasible under noise and target drift.

    \vspace{0.2em}


    \item \textbf{Disturbance-aware docking trajectory optimization through sequential convexification:}
    A finite-horizon SCP-based NMPC formulation is developed for aerial docking, incorporating moving-target prediction, convexified docking and collision constraints, and receding-horizon replanning to generate dynamically feasible docking trajectories under sensing uncertainty.
    
\end{itemize}

    

\section{System Dynamics}

Each quadrotor is simulated using full rigid-body dynamics within the MuJoCo physics engine. The high-fidelity simulation state is defined as
\begin{equation}
    \vec{x}^\top =
    \begin{bmatrix}
        \vec{p}^\top \; \vec{v}^\top \; \vec{q}^\top \; \vec{\Omega}^\top
    \end{bmatrix}
    \in \mathbb{R}^{13}
    \label{eq:quadrotor_state}
\end{equation}
where \( \vec{p} \in \mathbb{R}^3 \) and \( \vec{v} \in \mathbb{R}^3 \) denote position and linear velocity in the inertial frame, \( \vec{q} \in \mathbb{R}^4 \) represents the attitude quaternion, and \( \vec{\Omega} \in \mathbb{R}^3 \) is the body-frame angular velocity. This high-fidelity model is used only for simulation, while trajectory optimization employs a reduced-order model.

\subsection{Reduced-Order Model for NMPC}

The state used in the NMPC formulation is given by
\begin{equation}
\vec{x} =
\begin{bmatrix}
\vec{p}^{\top} & \vec{v}^{\top} & \phi & \theta & a_T & \vec{w}^{\top}
\end{bmatrix}^\top \in \mathbb{R}^{12}
\end{equation}
where $\phi$ and $\theta$ denote roll and pitch angles, $a_T$ represents the collective thrust acceleration, $\vec{p} \in \mathbb{R}^3$, $\vec{v} \in \mathbb{R}^3$ denote position and velocity in the inertial frame, and $\vec{w}$ represents the external wind disturbance included as a state.

The control input is defined as
\begin{equation}
\vec{u} =
\begin{bmatrix}
\phi_{\text{cmd}} & \theta_{\text{cmd}} & a_{\text{Tcmd}}
\end{bmatrix}^\top.
\end{equation}

Assuming negligible yaw dynamics and thrust aligned with the body vertical axis, the nonlinear dynamics used for trajectory optimization are given by
\begin{equation}
\begin{aligned}
\dot{\vec{p}} &= \vec{v}, 
&\quad \dot{v}_x &= a_T \sin(\theta), \\
\dot{v}_y &= -a_T \sin(\phi)\cos(\theta), 
&\quad \dot{v}_z &= a_T \cos(\phi)\cos(\theta) - g, \\
\dot{\phi} &= \frac{\phi_{\text{cmd}} - \phi}{\tau_{rp}}, 
&\quad \dot{\theta} &= \frac{\theta_{\text{cmd}} - \theta}{\tau_{rp}}, \\
\dot{a}_T &= \frac{a_{\text{cmd}} - a_T}{\tau_t}.
\end{aligned}
\end{equation}
where $\tau_{rp}$ and $\tau_t$ are actuator time constants.

This reduced formulation captures the dominant coupling between thrust direction and translational motion while avoiding the computational complexity of full rigid body dynamics. Wind disturbance is modeled as a stochastic random-walk process in the high-fidelity simulation and assumed piecewise constant over each NMPC prediction horizon.

Yaw dynamics are neglected under near hover or low speed forward flight conditions with approximately aligned chaser and target headings. Under this assumption, translational accelerations are dominated by roll and pitch induced thrust vectoring, while yaw has limited influence on the translational trajectory. Consequently, the model remains valid for small relative yaw angles and moderate angular rates, but does not capture heading misalignment effects or aggressive yaw-induced coupling.

\subsection{Linearisation}

At each SCP iteration, the nonlinear dynamics are linearised about the nominal trajectory 
$(\bar{\vec{x}}_k, \bar{\vec{u}}_k)$ as
\begin{equation}
f(\vec{x}_k,\vec{u}_k) \approx 
\bar{f}_k + \mathbf{A}_k(\vec{x}_k-\bar{\vec{x}}_k)
+ \mathbf{B}_k(\vec{u}_k-\bar{\vec{u}}_k),
\end{equation}
where $\bar{f}_k = f(\bar{\vec{x}}_k,\bar{\vec{u}}_k)$ and
\[
\mathbf{A}_k = \left.\frac{\partial f}{\partial x}\right|_{(\bar{x}_k,\bar{u}_k)}, 
\quad
\mathbf{B}_k = \left.\frac{\partial f}{\partial u}\right|_{(\bar{x}_k,\bar{u}_k)} .
\]

Using forward Euler discretization,
\begin{equation}
\vec{x}_{k+1} =
\vec{x}_k + \Delta t \Big(
\bar{f}_k
+ \mathbf{A}_k(\vec{x}_k-\bar{\vec{x}}_k)
+ \mathbf{B}_k(\vec{u}_k-\bar{\vec{u}}_k)
\Big) + \vec{\nu}_k ,
\end{equation}

where $\vec{\nu}_k$ is a virtual control introduced to maintain convex feasibility.

\subsection{Simulation Model and Control Architecture}

The reduced-order model is used only within the SCP optimization. The resulting control commands are applied to a full rigid-body simulation in MuJoCo governed by standard quadrotor dynamics:
\begin{equation}
\begin{aligned}
\dot{\vec{p}} &= \vec{v}, \\
m\dot{\vec{v}} &= m g \hat{e}_3 - \vec{f}\,\mathbf{R} \hat{b}_3, \\
\dot{\mathbf{R}} &= \mathbf{R} \tilde{\Omega}, \\
\mathbf{J}\dot{\vec{\Omega}} &= -\vec{\Omega} \times (\mathbf{J}\vec{\Omega}) + \vec{\tau}.
\end{aligned}
\end{equation}
where $\hat{b}_{3}$ denotes the body frame z axis, $\vec{f}$ is the total thrust, $\vec{\tau}$ is the control torque, $\textbf{J}$ is the body frame inertia matrix of the drone and $\Tilde{\Omega}(t)$ denotes the skew-symmetric form of the vector.

A hierarchical control architecture is employed:
\begin{itemize}
    \item The NMPC planner generates desired roll, pitch, and thrust commands $\vec{u}$ at a lower update rate.
    \item A high-frequency inner-loop PD controller stabilizes the attitude and tracks the commanded signals $(\phi_{\text{cmd}},\theta_{\text{cmd}},a_{\text{cmd}})$.
\end{itemize}

The resulting closed-loop system is nonlinear and disturbance-driven, and is iteratively approximated using SCP. This formulation enables efficient trajectory optimization while accounting for nonlinear rigid-body dynamics, actuator limits, and environmental disturbances.

\subsection{Target State Estimation}


The distressed target drone exhibits uncertain motion due to external disturbances, making future state prediction nontrivial. In this work, stationary and constant-velocity target motions are considered as baseline evaluation scenarios. Noisy position measurements from GPS or motion-capture systems are used for state estimation, while high-rate IMU measurements can further improve short-term prediction through sensor fusion. A discrete-time Kalman filter (KF)~\cite{ekf} based on a constant-velocity model is employed:


\begin{align}
\hat{\vec{x}}_{k+1|k} &= \mathbf{F}\hat{\vec{x}}_{k|k} + \vec{\eta}_k, \\
\vec{z}_k &= \mathbf{H}\vec{x}_k + \vec{\zeta}_k,
\end{align}
where $\mathbf{F}$ and $\mathbf{H}$ denote the state transition and measurement matrices, respectively. The process noise $\vec{\eta}_k$ and measurement noise $\vec{\zeta}_k$ are modeled as zero-mean Gaussian disturbances with covariances specified in Table~\ref{table:param}.

The filtered position and velocity estimates are propagated forward using the constant-velocity model to generate short-horizon predictions of the target motion. The predicted target state $\hat{\vec{x}}_t(t)$ is supplied to the SCP optimization routine at each planning step, enabling computation of relative dynamics and feasible docking trajectories despite measurement noise and disturbance-induced drift. This prediction mechanism allows the trajectory planner to anticipate target motion over the optimization horizon and maintain constraint-consistent interception behaviour. 

The proposed framework can be extended to accelerated or
disturbance-driven target maneuvers using higher-order target
prediction models.

\section{Methodology}

\begin{figure}
    \centering
    \includegraphics[width=0.5\textwidth]{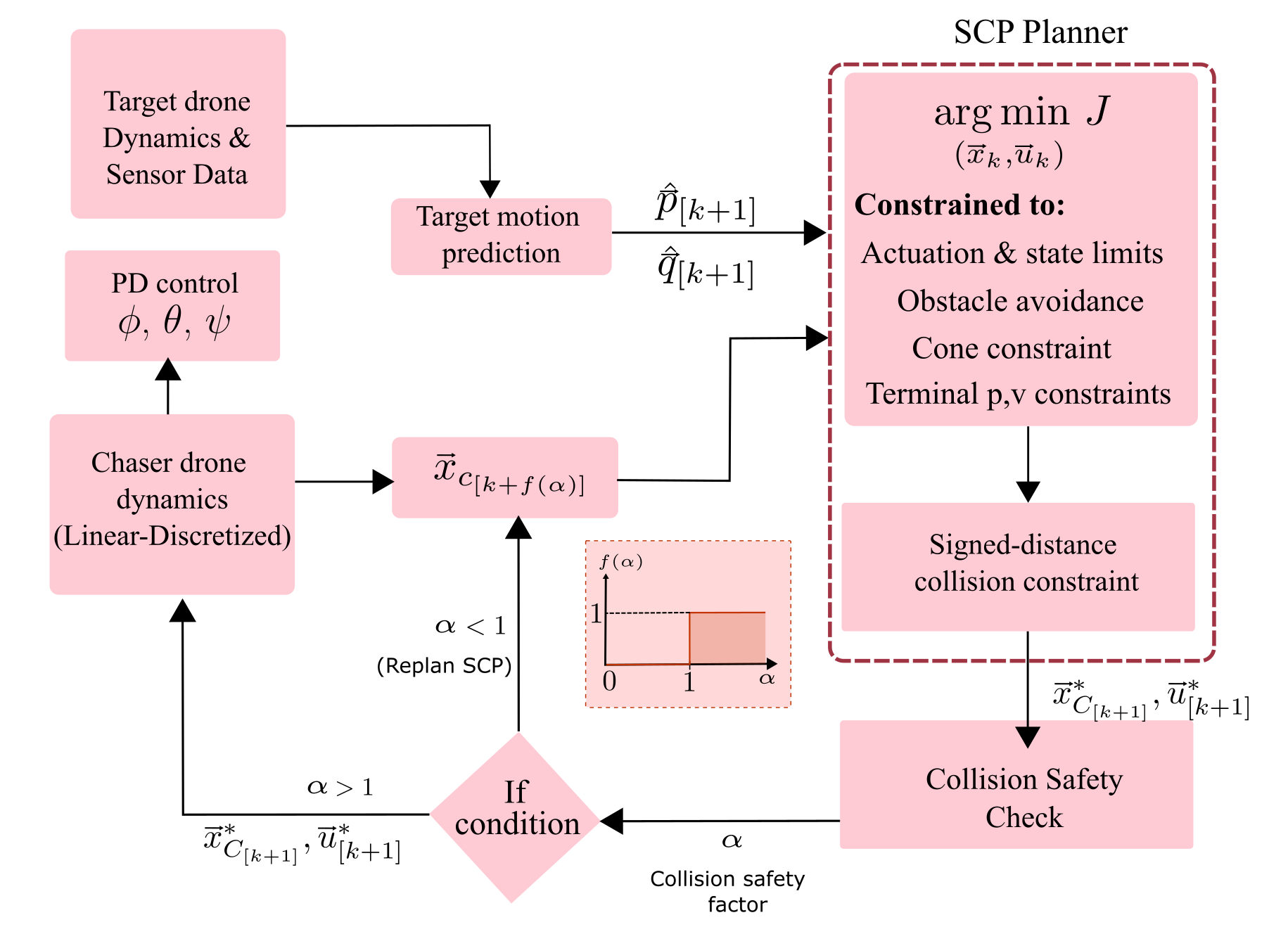}
    \caption{Flow chart of the SCP-based trajectory planning framework.}
    \label{fig:docking_flowchart}
\end{figure}

We formulate the docking maneuver as a minimal control effort trajectory optimization problem for a chaser quadrotor approaching a distressed target drifting with a low velocity. The system dynamics are discretised, and all nonconvex geometric and approach constraints are convexified into second-order cone constraints. This results in a sequence of convex subproblems formulated as a Second-Order Cone Program (SOCP) where finite-horizon optimal control problem is solved over a planning horizon of $N$ steps, enabling efficient real-time trajectory optimization \cite{sow2024convex}. 
Aerodynamic downwash is neglected in the present reduced-order formulation and will be investigated in future work through empirical characterization and higher-fidelity coupled aerodynamic models.

\subsection{Cost Function}



The SCP optimization problem minimizes a composite cost over the prediction horizon, expressed as sum of stage and terminal cost:

\begin{equation}
\min_{\{x_k,u_k,\nu_k,\sigma_k\}} 
\sum_{k=0}^{N-1} \ell(x_k,u_k,\nu_k,\sigma_k) 
+ \ell_f(x_N)
\end{equation}

\begin{equation}
\begin{aligned}
\ell(x_k,u_k,\nu_k,\sigma_k) &= 
\| \vec{p}_k - \vec{p}_{T,k} \|_Q^2
+ \| \vec{v}_k \|^2 
+ \| \boldsymbol{\phi}_k \|^2 \\
&\quad + \|u_k\|_R^2
+ \rho_u \|u_k - u_k^{nom}\|^2 \\
&\quad + \rho_\nu \|\nu_k\|^2
+ \rho_c \sigma_{\text{cone},k}
\end{aligned}
\end{equation}

\begin{equation}
\ell_f(x_N) =
\| \vec{p}_N - \vec{p}_{T,N} \|_{Q_f}^2
+ \| \vec{v}_N \|^2
+ \| \boldsymbol{\phi}_N \|^2
\end{equation}


where $\vec p_k,\vec v_k \in \mathbb{R}^3$, 
$\boldsymbol{\phi}_k=[\phi_k,\theta_k]^\top$ denote position, velocity,
and attitude states, and $u_k\in\mathbb{R}^3$ the control input. 
The weights are chosen as 
$Q = \mathrm{diag}(2,2,5)$, $Q_v=\mathbf I_3$, $Q_\phi=5\mathbf I_2$, 
$R=\mathrm{diag}(2,2,0.1)$, $Q_f=1000\mathbf I_3$, 
with terminal velocity and attitude penalties $=500$, and 
penalty coefficients $\rho_\nu=10^5$, $\rho_c=100$, $\rho_u=2$.

The cost function is defined as shown for the following reasons:
\begin{itemize}
\item \textbf{State and Control Effort: }  

The anisotropic weighting matrix $\textbf{Q} \succ 0$, used in the weighted Euclidean norm ($\rVert\cdot\rVert_{\mathbf{Q}}$), penalizes vertical tracking error (along the $\hat{e}_3$-axis) more than horizontal error. This encourages altitude alignment prior to lateral convergence during aggressive vertical maneuvers. 


\item \textbf{Soft Constraint Regularization:}
Additional penalty terms handle feasibility of the convexified dynamics and cone constraints through nonnegative slack variables. 


\end{itemize}

\subsection{Trust Region and Virtual Control}


To ensure the validity of the linear approximation used in SCP, a trust region constraint is imposed around the linearization trajectory. At each iteration, the state and control updates are restricted as
\begin{equation}
\|x_k - x_k^{\text{nom}}\|_\infty \leq r, \quad 
\|u_k - u_k^{\text{nom}}\|_\infty \leq r,
\end{equation}
where $r > 0$ is an adaptively updated trust-region radius. 

To maintain feasibility under linearization errors, a virtual control input $\nu_k$ is introduced in the dynamics. This modifies the discrete-time system as
\begin{equation}
x_{k+1} = x_k + \Delta t \, f_{\text{lin}}(x_k,u_k) + \nu_k,
\end{equation}
where $f_{\text{lin}}(\cdot)$ denotes the linearized dynamics.

The virtual-control and trust-region terms improve feasibility and numerical robustness under local linearization error. The penalty weight $\rho_\nu$ is chosen sufficiently large to drive $\nu_k \to 0$ at convergence, consistent with SCP regularization strategies such as GuSTO~\cite{gusto}.






\subsection{Actuation and State Limits}

Physical limitations of the quadrotor are enforced through convex input and state constraints. 
The commanded roll and pitch angles are bounded to respect tilt limits, while the thrust input is constrained within allowable actuation limits:
\begin{equation}
|u_{k,1}| \le \phi_{\max}, \quad 
|u_{k,2}| \le \theta_{\max}, \quad 
U_{\min} \le u_{k,3} \le U_{\max}.
\end{equation}

To ensure dynamically feasible motion and prevent aggressive maneuvers, the translational velocity is additionally constrained as
\begin{equation}
\|\vec{v}_k\|_2 \le V_{\max}.
\end{equation}

These constraints maintain physical realizability of the control inputs and promote safe trajectory generation within the quadrotor’s operational envelope.






\subsection{Docking Cone Constraint}

To ensure a physically admissible approach geometry for the docking mechanism, the chaser is constrained to remain within a conical capture region defined with respect to the target’s docking port. The cone constraint is activated when the relative lateral distance falls below a threshold, while a docking-axis offset guides the transition from standoff alignment to final capture.

\begin{equation}
\|\vec p_k - \vec p_T\|_2 \cos\theta_{\mathrm{dock}}
\le
-\vec n_{\mathrm{app}}^\top (\vec p_k - \vec p_T) + \sigma_{\mathrm{cone},k}
\label{eq:cone}
\end{equation}

where $\vec p_k$ denotes the chaser position at timestep $k$, $\vec p_T$ denotes the target docking interface position, $\vec n_{\mathrm{app}}$ is the docking axis direction, and $\theta_{\mathrm{dock}}$ is the half-angle of the docking cone. The slack variable $\sigma_{\mathrm{cone},k}\ge0$ is introduced to preserve feasibility of the convexified constraint.

\subsection{Collision and Obstacle Avoidance}

Collision avoidance between the chaser and the target drone is enforced using a signed distance constraint between their bodies. Each drone is modeled as a union of convex shapes, enabling efficient evaluation of pairwise signed distances as in \cite{dcol}. Let the poses of chaser and target at timestep $k$ be denoted as defined in \eqref{eq:signed}. The signed distance constraint is given as :

\begin{equation}
\alpha([\vec{p}_{C,k}, \vec{q}_{C,k}], [\vec{p}_{T,k}, \vec{q}_{T,k}]) \ge \alpha_{\min} - \sigma_{\alpha,k}
\label{eq:signed}
\end{equation}

which ensures a minimum separation $\alpha_{\min}$ and is enforced for all timesteps $k$.

Obstacle avoidance with static obstacles is handled using linearized convex constraints. For an obstacle with center $\vec{p}_{\mathrm{o}}$ and safety radius $R_{\mathrm{o}} + r_{\mathrm{safe}}$, the constraint is imposed as
\begin{equation}
\vec{n}_k^\top (\vec{p}_k - \vec{p}_{\mathrm{o}})
\ge
R_{\mathrm{o}} + r_{\mathrm{safe}}
\label{eq:collision}
\end{equation}
where $\vec{n}_k$ denotes the outward normal of the linearized obstacle boundary evaluated at the reference trajectory. Obstacle locations are assumed known and manually specified; perception-based detection is not considered in this study.





\subsection{Terminal Conditions}

Terminal capture is encouraged through a terminal cost evaluated at the end of the SCP prediction horizon. At the terminal step $k = N$, convergence of the chaser state toward the target docking interface is promoted by penalizing the relative position and velocity.

In simulation, successful capture is declared when
\begin{equation}
\begin{aligned}
\lVert \vec{p}_{C} - \vec{p}_{T} \rVert_2 &\le \varepsilon_p , \quad
\lVert \vec{v}_{C} \rVert_2 &\le \varepsilon_v
\end{aligned}
\end{equation}

where $\vec{p}_{C}, \vec{v}_{C}$ denote the chaser position and velocity, and $\vec{p}_{T}$ denotes the target docking interface position. Thus, terminal convergence is encouraged through cost shaping, while the docking condition itself is verified as a post-optimization feasibility check.

\section{Simulation Results}

The proposed approach was evaluated using a Python-based simulation framework on a PC equipped with an Intel i5-12400 CPU and 64~GB of RAM. The convex optimization subproblems generated by SCP were solved using the \textit{Clarabel} conic optimization solver. The simulation environment was implemented in \textit{MuJoCo}, with custom modules developed for trajectory planning and docking. The convex subproblems generated by SCP were solved in real time with an average solve time of approximately $12$–$18~\mathrm{ms}$ per iteration.

The proposed SCP-based docking framework is evaluated for stationary and constant-velocity target motions. Figures~\ref{fig:static} and~\ref{fig:linear} show representative docking trajectories for both scenarios, which were tested under identical initial conditions, obstacle configurations, controller parameters, and disturbance models. For evaluation, the target is modeled as a kinematic agent with stochastic disturbance inputs. This isolates the disturbance-rejection capability of the chaser while avoiding additional coupling from target control dynamics. The simulation study primarily evaluates the feasibility and robustness of the proposed docking framework under disturbance-driven conditions.

\begin{table}[H]
\caption{Simulation and NMPC Parameters}
\centering
\footnotesize
\setlength{\tabcolsep}{4pt}
\begin{tabular}{ll}
\hline
\textbf{Parameter} & \textbf{Value} \\
\hline
Chaser start $\vec p_C(0)$ & $[-2.5,\,0,\,1.5]^\top$ m \\
Target start $\vec p_T(0)$ & $[1.0,\,0,\,1.0]^\top$ m \\

Obstacle center $\vec p_{\mathrm{o}}$ & $[-1.0,\,0,\,1.25]^\top$ m \\
Obstacle radius $R_{\mathrm{o}}$ & $0.4$ m \\
Clearance radius $R_o + r_{\text{safe}}$ & $0.5$ m \\

Docking axis & $[0,\,0,\,-1]^\top$ \\
Docking cone angle $\theta_{dock}$ & $30^\circ$ \\

Thrust limits & $[2,\,15]$ m/s$^2$ \\
Max tilt angle & $25^\circ$ \\
Max velocity & $5$ m/s \\

Docking tolerances & $\epsilon_p=0.21$ m,\; $\epsilon_v=0.15$ m/s \\

    Wind disturbance model & $\dot{\vec w}\sim\mathcal N(0,0.1^2\mathbf I_3)$, $\|\vec w\|\le2.5$ m/s \\

EKF process covariance & diag$(0.05\mathbf I_3,\;0.2\mathbf I_3,\;0.5\mathbf I_3)$ \\

Planning horizon $N$ & $25$ \\
Time step $\Delta t$ & $0.1$ s \\
Max SCP iterations & $4$ \\


\hline
\multicolumn{2}{c}{\textbf{Target Model}} \\
\hline
Linear target velocity & $[0.15,\;0.10,\;0]^\top$ m/s \\
\hline
\end{tabular}
\label{table:param}
\end{table}

In all figures, the green curve and the red curve denote the planned and executed chaser trajectory respectively, the light-blue rays represent the docking cone defining admissible approach directions, the red translucent sphere indicates the obstacle region, and the purple and cyan hulls denote the target and chaser collision hulls, respectively. 


Simulation parameters are summarized in Table~\ref{table:param}. Unless otherwise stated, all reported performance metrics are averaged over multiple Monte-Carlo trials.

\subsection*{Case 1: Static Target}

\begin{figure}[H]
    \centering
    \includegraphics[width=1.0\linewidth]{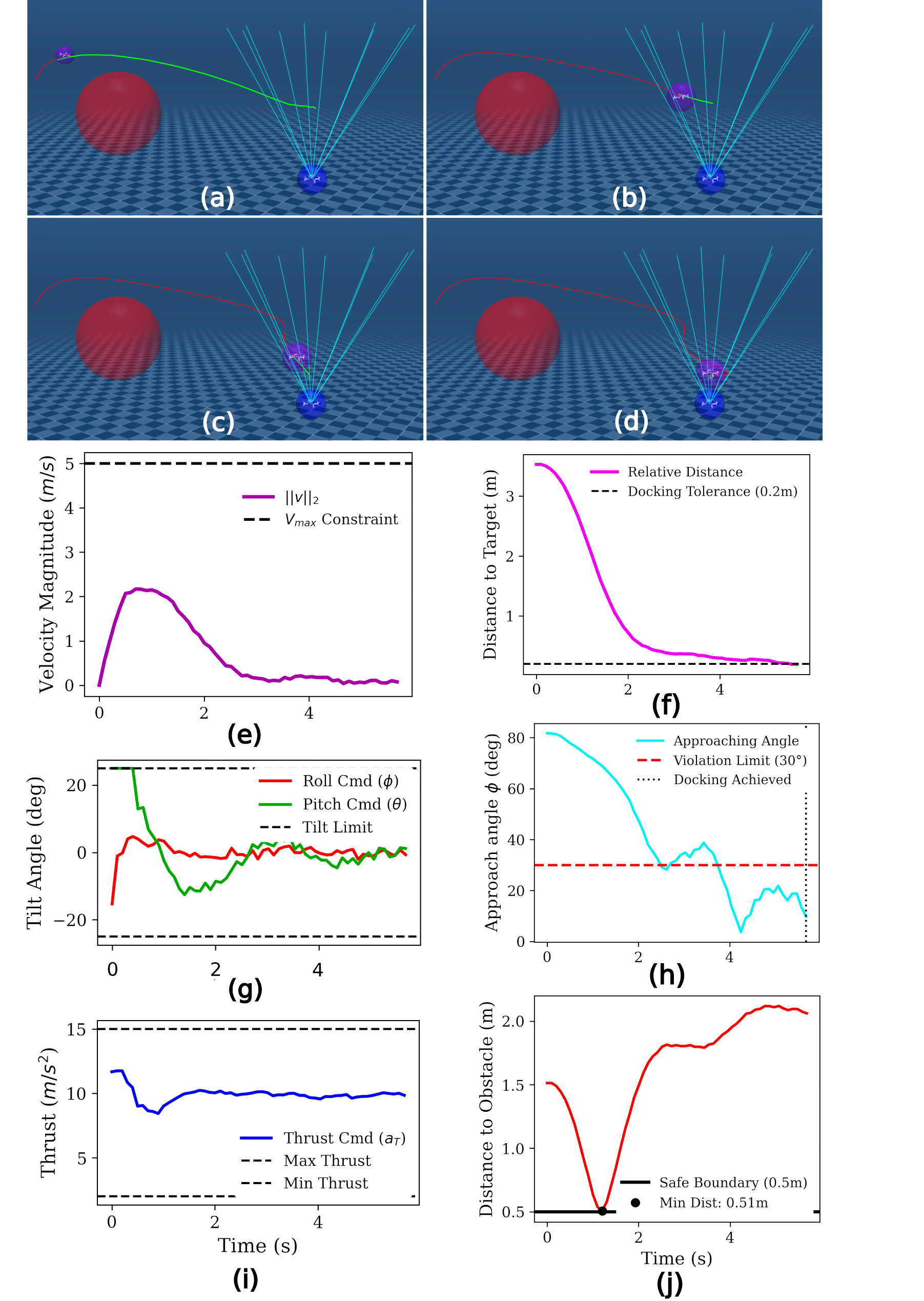}
    \caption{Case 1: \textbf{Static target}. Docking trajectory snapshots (a)-(d). Time histories of velocity magnitude (e), relative distance (f), tilt angles (g), approach angle and cone constraint (h), thrust (i), and obstacle clearance (j).}
    \label{fig:static}
\end{figure}

Figure~\ref{fig:static} shows the docking performance for a stationary target. 
The chaser follows a curved interception trajectory to satisfy obstacle avoidance 
and cone alignment constraints. The velocity peaks at approximately $2.2~\mathrm{m/s}$ 
before decreasing during the constrained approach phase. Tilt commands briefly 
approach actuator limits during the initial maneuver and subsequently stabilize 
as alignment with the docking axis is achieved. The approach angle enters the admissible cone region after $\approx3~\mathrm{s}$, 
while the minimum obstacle clearance remains above the safety threshold 
($\approx0.5~\mathrm{m}$). Docking is achieved at approximately $6~\mathrm{s}$, 
with the relative position and velocity converging within the prescribed tolerances.

\subsection*{Case 2: Linearly Moving Target}

\begin{figure}[H]
    \centering
    \includegraphics[width=1.0\linewidth]{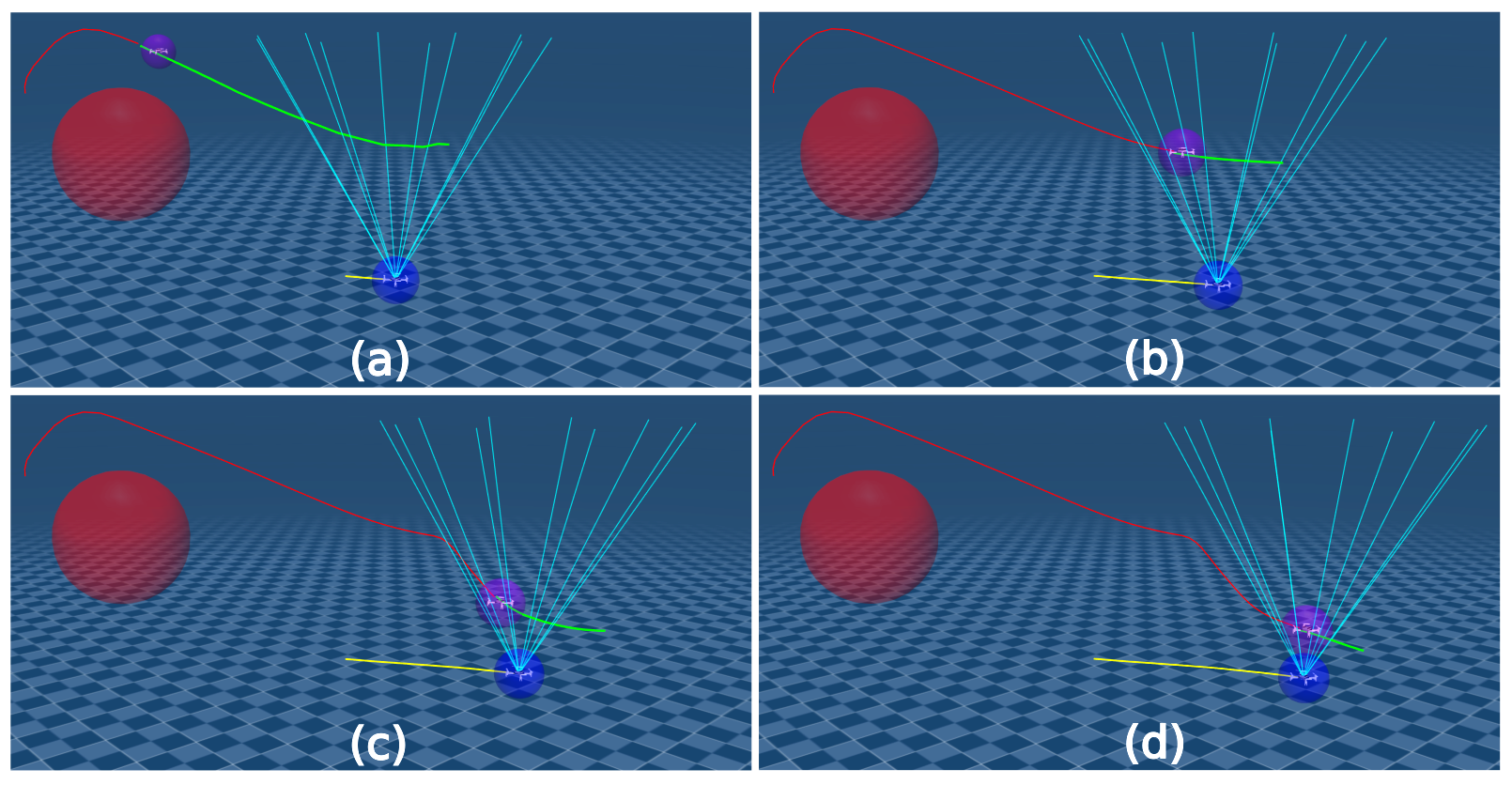}
\caption{Case 2: \textbf{Linear target motion}. Docking trajectory snapshots (a)–(d).}
    \label{fig:linear}
\end{figure}

For a target undergoing constant linear motion, interception requires sustained corrective maneuvers, leading to a longer convergence time. Representative docking snapshots from the MuJoCo simulation are shown in Fig.~\ref{fig:linear}, illustrating the evolving interception geometry.


\begin{figure}[t]
    \centering
    \includegraphics[width=0.8\linewidth]{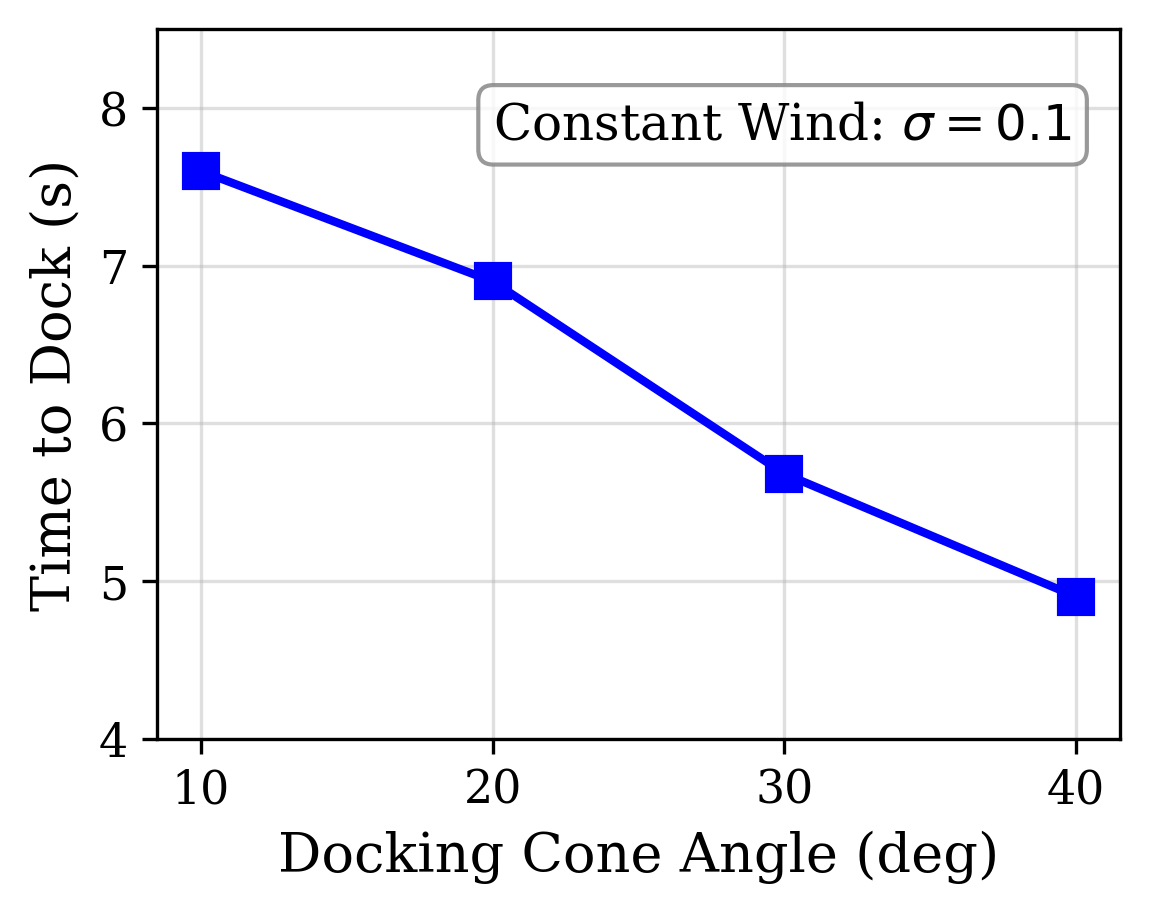}
    \caption{Variation of docking time (s) with the approach cone angle (deg)}
    \label{fig:cone_expt}
\end{figure}

Figure~\ref{fig:cone_expt} illustrates the effect of docking cone angle on docking time under constant wind disturbance ($\sigma=0.1$). Larger cone angles reduce convergence time by relaxing approach constraints and vice versa.

\begin{table}[H]
\centering
\caption{Docking Performance under Wind Disturbances [10 Trials]}
\begin{tabular}{c c c c}
\hline
\textbf{Wind $\sigma$} & 
\textbf{Dock Time (s)} & 
\textbf{\shortstack{Peak Cone\\Violation ($^\circ$)}} & 
\textbf{\shortstack{Control\\Energy $E$}} \\
\hline
0.1 & 5.68 & 0.0 & 1.24 \\
0.3 & 5.98 & 0.2 & 1.39 \\
0.5 & 6.70 & 0.7 & 1.65 \\
1.0 & $>15$ (fail) & 5.8 & 2.87 \\
\hline
\end{tabular}
\vspace{2mm}

{\footnotesize $E = \sum_k \|u_k\|^2 \Delta t$. Peak cone violation denotes the maximum instantaneous angular violation.}
\label{tab:wind_docking}
\end{table}

Table~II reports docking performance under varying wind disturbance levels. 
Increasing disturbance intensity leads to longer convergence times and higher 
control effort. For large disturbance levels ($\sigma > 0.8$), the docking task becomes infeasible within the finite prediction horizon, primarily due to repeated violation of the docking-cone and
terminal convergence conditions. Increased target drift causes larger corrective maneuvers and higher control effort, eventually leading to loss of consistent approach alignment. Similar
sensitivity to relative-motion uncertainty has also been observed in convex docking formulations such as~\cite{sow2024convex}.

\section{Conclusion}


This work presents a convex trajectory optimization framework based on Sequential Convex Programming for autonomous mid-air drone-to-drone docking under dynamic target motion and environmental disturbances. The results demonstrate stable docking behaviour across multiple target motion scenarios while maintaining bounded control effort and consistent constraint satisfaction, indicating the suitability of convex optimization for real-time aerial docking tasks.


Future work will focus on deploying the framework on a physical quadrotor platform in a constrained indoor environment to validate onboard computation and sensing, including the effects of multirotor downwash interactions. Additional research directions include incorporating energy-aware cost formulations and notions of optimality, extending the framework to include full orientation control of the chaser, and developing a slow and robust terminal capture phase. Future work will also investigate formal recursive feasibility and closed-loop stability analysis, more realistic target behaviors including accelerated and aggressively maneuvering aerial targets, and adaptive target prediction within the receding-horizon framework. A systematic comparison with classical controllers (e.g., PID) and alternative nonlinear optimization methods such as SQP will additionally be conducted to assess performance in constraint satisfaction, control effort, and computational scalability.

\section*{Acknowledgment}

The authors thank Yashwanth M. and Harishankar M. for their valuable suggestions and discussions.

\bibliographystyle{IEEEtran}
\bibliography{ref}

\balance
\end{document}